\documentclass[11pt]{article}

\usepackage[final]{acl}

\usepackage{times}
\usepackage{latexsym}
\usepackage[textsize=small]{todonotes}
\usepackage[T1]{fontenc}

\usepackage[utf8]{inputenc}

\usepackage{microtype}
\usepackage{amsfonts}
\usepackage{xspace}
\usepackage{inconsolata}
\usepackage{amsmath} 
\usepackage{graphicx}
\usepackage{algorithm}
\usepackage{algpseudocode}
\usepackage{amsmath,amssymb} 
\usepackage{amsthm} 
\usepackage{cleveref}

\usepackage[dvipsnames]{xcolor}
\definecolor{darkestgreen}{rgb}{0.0, 0.5, 0.0}
\usepackage{booktabs}
\usepackage{float}

\newtheorem{assumption}{Assumption}
\newtheorem{lemma}{Lemma}
\newtheorem{theorem}{Theorem}
\newtheorem{corollary}{Corollary}
\newtheorem{remark}{Remark}

\DeclareMathOperator{\DepErr}{DepErr}
\newcommand{\modelname}{\textbf{{MEDAL}}\xspace}

\title{Diffusion Language Model Inference with Monte Carlo Tree Search}

\usepackage{hyperref}

\author{
Zheng Huang$^{1,2} \thanks{Work done while interning at AWS AI Labs.$^{\dagger}${Correspondence:} \href{zheng.huang.gr@dartmouth.edu}{zheng.huang.gr@dartmouth.edu}, 
\href{raxkiran@amazon.com}{raxkiran@amazon.com}, \href{yyanc@amazon.com}{yyanc@amazon.com}.}$ \quad
Kiran Ramnath$^{1,\dagger}$ \quad
Yueyan Chen$^{1,\dagger}$ \quad 
Aosong Feng$^{1}$ \quad \\
\textbf{Sangmin Woo$^{1}$} \quad
\textbf{Balasubramaniam Srinivasan$^{1}$} \quad 
\textbf{Zhichao Xu$^{1}$} \quad \\
\textbf{Kang Zhou$^{1}$} \quad
\textbf{Shuai Wang$^{1}$} \quad
\textbf{Haibo Ding$^{1}$} \quad
\textbf{Lin Lee Cheong$^{1}$} \quad
\\
$^{1}$AWS AI Labs,
$^{2}$Dartmouth College\\
}

\begin{document}

\maketitle
\begin{abstract}
Diffusion language models (DLMs) have recently emerged as a compelling alternative to autoregressive generation, offering parallel generation and improved global coherence. During inference, DLMs generate text by iteratively denoising masked sequences in parallel; however, determining which positions to unmask and which tokens to commit forms a large combinatorial search problem. Existing inference methods approximate this search using heuristics, which often yield suboptimal decoding paths; other approaches instead rely on additional training to guide token selection. To introduce a principled search mechanism for DLMs inference, we introduce \modelname, an inference-time scaling framework that integrates \textbf{M}onte Carlo Tree S\textbf{E}arch initialization for \textbf{D}iffusion L\textbf{A}nguage Mode\textbf{l} inference. We employ Monte Carlo Tree Search at the initialization stage to explore promising unmasking trajectories, providing a robust starting point for subsequent refinement. This design enables efficient inference-time scaling, allowing generation quality to improve as the search budget increases, without additional training.
Across multiple benchmarks, \modelname achieves up to 22.0\% improvement over existing inference strategies, establishing a new paradigm for search-based inference in DLMs.

\end{abstract}

\section{Introduction}
In recent years, diffusion language models (DLMs) have emerged as a powerful alternative for generative modeling over discrete sequences \cite{zhu2025llada, ye2025dream}. Unlike autoregressive (AR) models~\cite{achiam2023gpt, minaee2024large}, which rely on a strictly left-to-right factorization, DLMs learn to invert a stochastic corruption process that independently masks tokens \cite{lou2023discrete, shi2024simplified, nie2025large}. This formulation enables parallel refinement, improves global coherence, and offers flexible quality-latency trade-offs, thus challenging the dominance of AR paradigms \cite{li2025survey}.

The inference process of DLMs can be naturally formulated as a search problem: starting from a corrupted sequence, the model iteratively decides which positions to unmask and which tokens to assign, navigating an exponential space of possible trajectories. Existing inference-time methods approximate this search using confidence-driven heuristics, such as greedily unmasking the highest-confidence tokens~\cite{kim2025train, ben2025accelerated, luxembourg2025plan}. Although effective in reducing short-term uncertainty, these strategies are inherently myopic: once high-confidence tokens are fixed, subsequent steps are forced to adapt around them, often leading to suboptimal trajectories. Another line of work adjusts masking schedules dynamically~\cite{peng2025path, zhao2024informed}, but such approaches typically require training auxiliary samplers to determine token updates at each step, introducing additional complexity and limiting general applicability.

These limitations highlight the need for a principled search mechanism that can explore alternative unmasking trajectories without additional training overhead. To this end, we propose \modelname, a framework that integrates \textbf{M}onte Carlo Tree S\textbf{E}arch initialization for \textbf{D}iffusion L\textbf{A}nguage Mode\textbf{l} inference. Unlike heuristic or schedule-based methods, which either commit to tokens greedily or depend on auxiliary samplers, \modelname adopts a principled search-based approach, using Monte Carlo Tree Search (MCTS)~\cite{yoon2025monte, browne2012survey} to balance exploitation of high-confidence tokens with exploration of alternative unmasking trajectories relying on signals obtained from the model's own distribution. We propose two key innovations that enable MCTS to be applied effectively to DLM inference. First, we introduce a confidence-guided filtering mechanism that focuses inference on the most promising tokens and positions. Second, we design an information-gain reward that guides MCTS by favoring token choices that not only resolve the current position but also increase the model’s confidence in predicting the remaining tokens. Together, these innovations make it possible to enable the four stages of MCTS, Selection, Expansion, Simulation, and Backpropagation, within DLMs.
Rather than applying MCTS exhaustively, we employ it strategically in the early stages of inference to construct a robust initialization, after which the process continues with efficient heuristics. To further address complex prompts that induce high uncertainty, we incorporate a task-decomposition module that automatically splits the input into smaller subtasks, thereby reducing ambiguity and providing structured guidance for subsequent unmasking decisions.
Our contributions are summarized as follows:
\begin{itemize}
    \item \textbf{Novel Formulation}: We frame DLM inference as a search problem and introduce \modelname, the first framework to integrate MCTS into DLM inference, enabling principled exploration beyond greedy heuristics or schedule-based methods
    \item \textbf{Novel Design}: We design a new DLM inference approach that combines an MCTS-guided initialization module with a task-decomposition module, enabling both efficient search and improved handling of complex tasks.
    \item \textbf{Extensive Experiments}: We conduct evaluations on various benchmarks, demonstrating that our method outperforms existing inference strategies for DLMs by up to 22.0\% when restricting MCTS to initialization. We show that generation quality continues to improve even further with diminishing gains as the MCTS initialization budget increases, validating the effectiveness of our search-based approach. 
\end{itemize}

\section{Preliminary}

\begin{figure*}[t]
    \centering
    \includegraphics[width=\textwidth]{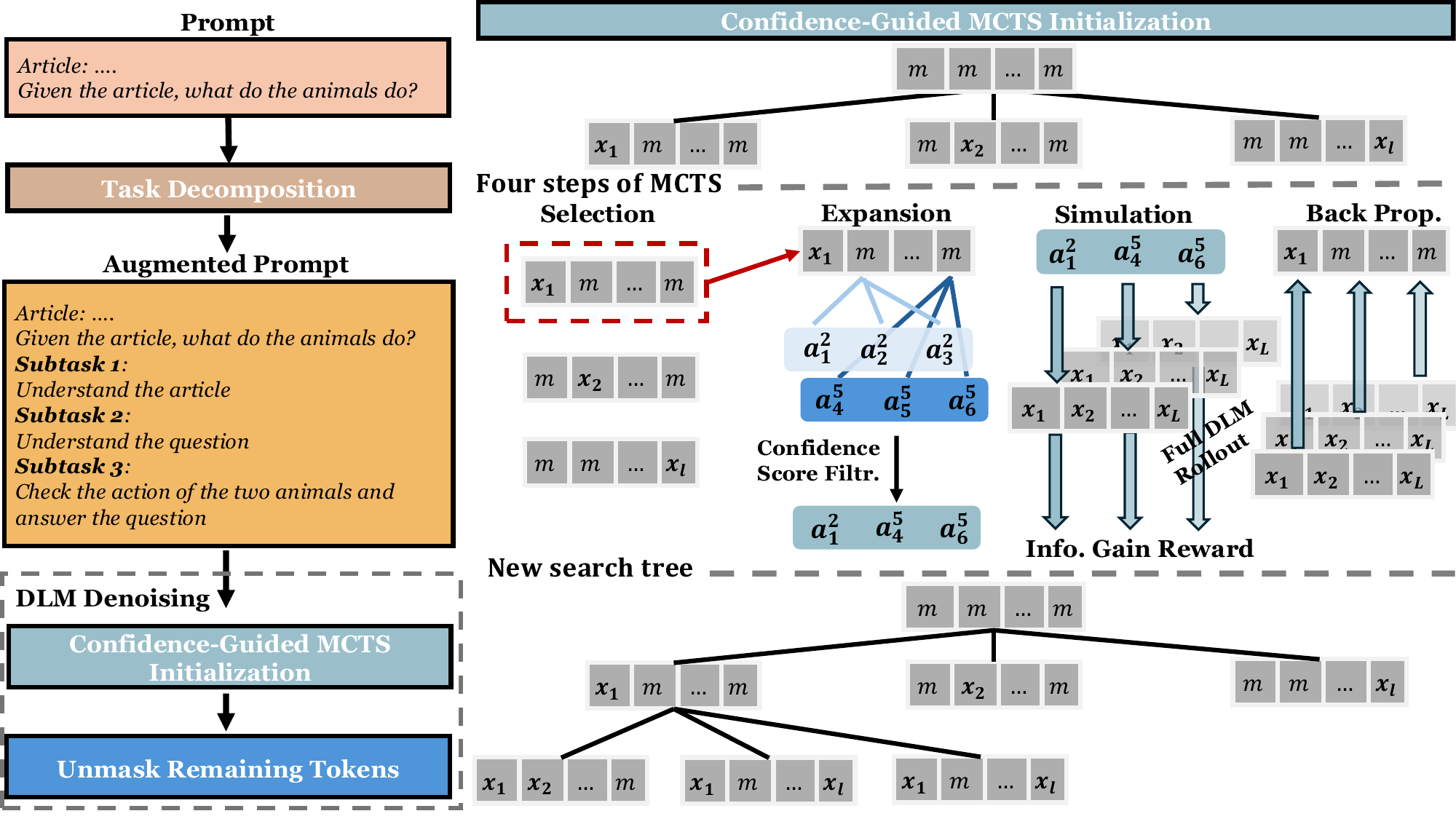}
    \caption{Overview of the \modelname framework. It consists of an MCTS-guided initialization module and a task-decomposition module, enabling both efficient search and improved handling of complex tasks. The root node represents the initial fully masked input, and leaf nodes correspond to partially unmasked sequences. The MCTS involves committing to a specific generation decision, expressed as an action $a_v^i = (i, v)$, where $i$ identifies the masked position to reveal and $v$ denotes the token selected for that position.}
    \label{fig:framework}
\end{figure*}

\subsection{Discrete Diffusion Language Models}

DLMs adapt the diffusion paradigm from continuous domains (e.g., image generation) to discrete text sequences. Let $x_0 = (x^1_0, \dots, x^L_0)$ denote a token sequence of length $L$ sampled from the data distribution. The core idea is to define a forward noising process that progressively corrupts $x_0$ into increasingly noisy sequences $\{x_t\}_{t=1}^T$, and to train a neural model to learn the corresponding reverse denoising process that reconstructs $x_0$ from noise.

\paragraph{Forward process.}
In discrete DLMs, the forward process is typically defined by a time-dependent transition matrix $Q_t$ over the vocabulary \cite{li2025survey}. At each time $t$, the probability of a
state $x_t$ given an initial state $x_0$ is given by a categorical
distribution: 
\begin{equation}
    q(x_t \mid x_0) = \text{Cat}(x_t; x_0 \overline{Q}_t), \quad \overline{Q}_t = \prod_{i=1}^t Q_i.
\end{equation}


\paragraph{Reverse Process}
The reverse process learns to invert the corruption by predicting the original token distribution given a corrupted sequence:
\begin{equation}
    p_\theta(x_{t-1} \mid x_t) = \prod_{{i \in \mathcal{U}_t}} p_\theta(x^i_{t-1} \mid x_t, t).
\end{equation}
Here, $p_\theta$ is parameterized by a transformer, trained to minimize a cross-entropy objective over masked positions, and $\mathcal{U}_t$ denotes the set of positions unmasked at timestep $t$.

\paragraph{Generation.}
During inference, generation begins from a fully masked sequence:
\begin{align}
    x_T = (\texttt{[MASK]}, \dots, \texttt{[MASK]}).
\end{align}
At each denoising step $t$, the model outputs a distribution over the vocabulary for every position. A subset of tokens with the highest confidence is selected, unmasked, and fixed, while the remaining positions stay masked. The process then advances to step $t-1$, where the model re-predicts distributions conditioned on both the fixed tokens and the still-masked positions~\cite{zhu2025llada, nie2025large}. This iterative refinement continues until all positions are resolved, yielding the final sequence $x_0$. The generation can be viewed as a sequence of partially completed states, where the model progressively transitions from a fully masked input to a coherent, fully unmasked output.

\subsection{Monte Carlo Tree Search}
MCTS is a general algorithm for decision-making in large combinatorial search spaces, where exhaustive enumeration of all trajectories is infeasible~\cite{browne2012survey}. The goal of MCTS is to evaluate possible decision paths efficiently, balancing the exploration of new options with the refinement of promising ones. The search tree begins at the root node, representing the initial state of the problem, and grows toward leaf nodes, which correspond to unexplored frontier states. Each iteration of MCTS consists of four steps: 
(i) Selection: choose the node from the root to a leaf according to a selection policy, such as the upper confidence bound (UCB);
(ii) Expansion: add one or more child nodes at the leaf;
(iii) Simulation: evaluate the newly expanded node by performing rollouts or applying heuristic approximations;
(iv) Back propagation: propagate the result upward to update statistics of visited nodes.

\section{Method}

In this section, we present our framework \modelname for enhancing DLM inference (Figure~\ref{fig:framework}). We start with the notation used throughout the section. Second, we introduce our key contributions: confidence score filtering (\Cref{sec:conf_filter}) and information-gain reward (\Cref{sec:infog}) that enable MCTS to be effectively applied to DLMs. Building on these foundations, we then introduce our MCTS-guided initialization strategy to efficiently explore promising candidates during the early stages of generation. Finally, we describe how task decomposition via prompt guidance can further reduce uncertainty and improve generation quality.

\paragraph{Notation. }
We use $\mathcal{V}$ for the vocabulary (size $|\mathcal{V}|$); $L$ for target sequence length; $x=(x^{1},\dots,x^{L})$ for a (partially) formed sequence; $x^{i}$ the token at position $i$; $x_{\setminus i}$ the sequence with position $i$ masked; $t\in\{0,\dots,T\}$ the reverse denoising step index with $T$ total steps; $\mathcal{M}_{t}$ the set of masked positions at step $t$.

\subsection{MCTS for DLMs Inference}
To bridge MCTS with DLMs, we construct the search tree over unmasked sequences, where the root corresponds to the initial masked input, and leaves as partially unmasked sequences.

\subsubsection{Confidence Score Filtering}
\label{sec:conf_filter}
 The classical MCTS aims to balance exploitation and exploration, but in DLMs the search space spans the full vocabulary at every masked position, making naïve search intractable. To address this, we introduce a confidence filtering strategy that restricts the search space of masked positions and tokens to a far smaller action set. We define the action as a position–token pair $a_v^i=(i, v)$, where $i$ is a masked position and $v$ is the token-candidate chosen, and it represents a specific unmasking decision of the search precess. Next, we describe how to build the action set.

Given the model's logits $\ell^i \in \mathbb{R}^{|\mathcal{V}|}$ of an input sequence at a masked position $i$,
we first convert $\ell^i $ into probabilities
\begin{align}
p^i(v) = \frac{\exp(\ell^i_v)}{\sum_{w \in \mathcal{V}} \exp(\ell^i_w)}, \qquad v^i \in \mathcal{V}.
\end{align}

For a candidate token $v$ at position $i$, we then define the confidence-adjusted score as:
\begin{equation} 
s^i_v = p^i(v) \cdot \phi_{\text{ent}}^i \cdot \phi_{\text{mar}}^i,
\end{equation}

where $\phi_{\text{ent}}^i$ and $\phi_{\text{mar}}^i$ are respectively the entropy penalty and top-2 margin, defined below: 
\begin{itemize}
    \item \textbf{Entropy Penalty.} We use
    $H^i = - \sum_{v \in \mathcal{V}} p^i(v)\log (p^i(v) + \varepsilon)$, with $\phi_{\mathrm{ent}}^i = \exp(-H^i) \in (0,1]$, where $\varepsilon$ is a small constant ensuring numerical stability.
    Positions with higher entropy (low confidence) are down-weighted.  
    \item \textbf{Top-2 Margin.} 
    Let $p^i_{(1)} \geq p^i_{(2)}$ be the top-two probabilities. The margin $\Delta^i = p^i_{(1)} - p^i_{(2)}$ is mapped to $\phi_{\mathrm{mar}}^i = \sigma(\gamma \Delta^i)$ with $\gamma$ is a hyperparameter and $\sigma$ is a sigmoid function.
\end{itemize}

For each masked position $i \in \mathcal{M}_t$, we retain the top-$K_1$ candidate tokens ranked by the confidence-adjusted score $s^i_v$, forming an action candidate set:

\begin{align}
    \mathcal{A}^i_t &= \{a_v^i: v \in \text{Top-}K_1(s^i_v)\} \\
\mathcal{A}_t &= \bigcup_{i \in \mathcal{M}_t} \mathcal{A}^i_t.
\end{align}

From the action candidate set $\mathcal{A}_t$, we then select top-$K_2$ actions with the highest score:
\begin{align}
    \mathcal{A}_t^\prime = arg \text{ Top-}K_2\{s_v^i: a_v^i \in \mathcal{A}_t\} 
\end{align}

The best $K_2$-scoring actions are then applied to the simulation of MCTS (\Cref{sec:mcts_steps}).




%


\subsubsection{Information-Gain Reward} 
\label{sec:infog}

\begin{algorithm}[t]
\caption{Confidence-Guided MCTS for DLMs Inference (CGMCTS)} 
\label{alg:mcts_dlm}
\begin{algorithmic}[1]
\Require Prompt tokens $x_{1:|P|}$, target length $L$, MCTS init length $L_c{<}L$, action candidate token number $K_1$, the highest-scoring number $K_2$, candidate size $C$, margin scale $\gamma$. 
\State Initialize root$ \gets [x_{1:|P|}] \oplus [\texttt{[MASK]}]^{L}$ 
\While{(CollectCandidates(root, C, $L_c$))} \Comment{Stop when collect $C$ candidates with length $L_c$}
    \State $x \gets $ NodeSelection($root$) \Comment{UCB selection}
    \State Identify masked positions $\mathcal{M}_t$ in $x$
    \ForAll{$i \in \mathcal{M}_t$}
        \State Compute distribution $p^i(v) = \mathrm{softmax}(\ell^i)$
        \State Compute confidence factors:
        $\phi_{\mathrm{ent}}^i = \exp(-H^i)$,
        $\phi_{\mathrm{mar}}^i = \sigma(\gamma \Delta^i)$
        \State Score tokens:
        $s^i_v = p^i(v) \cdot \phi_{\mathrm{ent}}^i \cdot \phi_{\mathrm{mar}}^i$
        \State Select candidates $\mathcal{A}^i_t \gets \mathrm{Top}\!-\!K_1(s^i_v)$
    \EndFor
    \State Aggregate $\mathcal{A}_t = \bigcup_{i}\mathcal{A}^i_t$
    \State Build $\mathcal{A}_t^\prime=arg \text{ Top-}K_2\{s_v^i: a_v^i \in \mathcal{A}_t\} $
    
    \ForAll{$a \in \mathcal{A}_t^\prime$}
    \State $x_a \gets$ Apply action $a$ to $x$
    \State Node$ \gets \text{ExpandNode}(X_a)$ 

    \State Rollout to obtain $r_{IG}(a)$
    \State Backpropagate $r_{IG}(a)$ \Comment{Update search tree}

    \EndFor
    
\EndWhile

\State \Return $C$ collected candidates with length $L_c$
\end{algorithmic}
\end{algorithm}

Given the selected action set $\mathcal{A}_t'$, the next step is to evaluate the impact of each candidate action. 
An effective action should not only specify a token at one masked position, but also provide contextual information that improves the model’s confidence over the remaining masked positions. 
In other words, a good choice at position $i$ should make the subsequent predictions for $\mathcal{M}_t \setminus \{i\}$ more confident.

Our objective is therefore to design a reward function that quantifies how much an action improves the model’s confidence in the unresolved positions. 
Formally, for an action $a_v^i \in \mathcal{A}_t'$, selecting token $v$ to fill position $i$ updates the masked set from $\mathcal{M}_t$ to $\mathcal{M}'_t = \mathcal{M}_t \setminus \{i\}$. We define the reward as the information-gain reward, which measures the entropy reduction across the remaining positions:
\begin{equation}
\resizebox{0.9\columnwidth}{!}{$
r_{\mathrm{IG}}(a_v^i) \;=\; 
\frac{\sum_{j \in \mathcal{M}_t} H^{\text{before}}_\theta(j) 
\;-\; \sum_{j \in \mathcal{M}'_t} H^{\text{after}}_\theta(j)}{\sum_{j \in \mathcal{M}_t} H^{\text{before}}_\theta(j)},
$}
\label{info_gain}
\end{equation}

where $H_\theta(j)$ denotes the predictive entropy at position $j$. 
Position $i$ is excluded from the second summation: once it has been filled, its entropy is always zero regardless of the token chosen, and including it would contribute only a constant term. 

Finally, each action in $\mathcal{A}_t'$ is rolled out and assigned a reward value, which serves as the feedback signal to guide tree selection. This design ensures that the search prioritizes actions that not only fill a mask but also help improve model confidence for future predictions.

\subsubsection{MCTS Steps}
\label{sec:mcts_steps}
Building on the description above, we detail how the four
traditional steps of MCTS—Selection, Expansion, Simulation, and Backpropagation—are adapted. The algorithm overview is in Algorithm~\ref{alg:mcts_dlm}, and an example is shown in~\Cref{app_mcts_example}. 

\paragraph{Selection.}

In this phase, we traverse the tree from the root to a leaf using UCB~\cite{yoon2025monte, kocsis2006bandit}, selecting the child node that maximizes $Q(x, a) + c\sqrt{\frac{\ln N(x)}{N(x,a)}}$, where $x$ represents the current partially masked sequence state, $Q(x, a)$ is the mean reward of action $a$, $N(x)$ is the visit count of node $x$, $N(x, a)$ is the visit count of the child node, and $c$ is the exploration constant.

\paragraph{Expansion}
We build the action candidate set $\mathcal{A}_t'$ using confidence score filtering (\Cref{sec:conf_filter}). We then pass each action $a_v^i \in \mathcal{A}_t'$ to the simulation step to evaluate its reward.

\paragraph{Simulation.}
The simulation step evaluates the impact of each action $a_v^i \in \mathcal{A}_t'$ by rolling out the remaining masked positions by DLMs. Specifically, after applying action $a_v^i$ to fill position $i$ with token $v$, we let the DLM fill the remaining masked positions $\mathcal{M}'_t = \mathcal{M}_t \setminus \{i\}$ by sampling from the model's predicted distribution. This yields a completed sequence. The information-gain reward $r_{IG}(a_v^i)$ is then computed based on the entropy reduction across the remaining masked positions, as described in \Cref{sec:infog}.

\paragraph{Backpropagation.}

After the simulation step, the reward
obtained from evaluating the complete sequence is backpropagated through the tree to update the value estimates of
all parent nodes along the path to the root.

\begin{algorithm}[!t]
\caption{Overview of \modelname}
\label{alg:td-umcts}
\begin{algorithmic}[1]
\Require Prompt tokens $x_{1:|P|}$, target length $L$, total steps $T$, MCTS init length $L_c{<}L$, action candidate token number $K_1$, the highest-scoring number $K_2$, candidate size $C$, margin scale $\gamma$. 
\Ensure Completed sequence $x$
\Statex \textbf{(A) Task decomposition}
\State $\hat{P} \gets \Call{TaskDecompose}{x_{1:|P|}}$ \Comment{augment prompt with self-generated subtasks}
\State $x \gets [\hat{P}] \oplus [\texttt{[MASK]}]^L$
\Statex
\Statex \textbf{(B) MCTS-based initialization}
\State Candidates$ \gets \text{CGMCTS}(x, L, L_c, K_1, K_2, \gamma)$\label{line:call-umcts}

\State $x \gets $ SelectCandidate(Candidates) \Comment{Select the one that has the highest info. gain reward}

\Statex
\Statex \textbf{(C) Confidence-guided unmasking for remaining steps}
\For{$t = T$ to $1$}
  \State Identify masked positions $\mathcal{M}_t$ in $x$; \textbf{if} $\mathcal{M}_t{=}\varnothing$ \textbf{ then break}
  \ForAll{$i \in \mathcal{M}_t$}
      \State Compute distribution $p^i(v)=\mathrm{softmax}(\ell^i)$
      \State Compute confidence factors $\phi_{\mathrm{ent}}^i=\exp(-H^i)$,\quad $\phi_{\mathrm{mar}}^i=\sigma(\gamma \Delta^i)$
      \State Score tokens $s^i_v = p^i(v)\cdot \phi_{\mathrm{ent}}^i \cdot \phi_{\mathrm{mar}}^i$
  \EndFor
  \State Sample action $a^i_v\propto \mathrm{Softmax}(s^i_v)$
  \State Apply action: $x[i] \gets v$
\EndFor
\State \Return $x$
\end{algorithmic}
\end{algorithm}

\subsection{Confidence-Guided MCTS Initialization}
While conceptually appealing, applying MCTS throughout the entire generation is computationally prohibitive: the branching factor scales with vocabulary size $|\mathcal{V}|$ and sequence length $L$, and repeated rollouts over $T$ denoising steps incur exponential cost.
This leads to full-sequence MCTS being impractical for large-scale DLM inference. To balance search quality with efficiency, we restrict MCTS to the initialization phase. Specifically, we terminate the MCTS once a candidate set of $C$ partially unmasked sequences, each of length $L_c$, has been formed. Each candidate is then fully rolled out by the DLM to compute its information-gain reward, and the sequence with the highest reward is selected as the final resolved output. Once a partially resolved sequence is obtained, the remaining tokens are filled without further tree search: at each step, we directly apply the confidence-adjusted score to select high-confidence tokens until no masks remain.  
This strategy enables test-time scaling during critical early decision stages through structured search over generation trajectories, while maintaining tractable inference via confidence-guided decoding in later stages.

\subsection{Task Decomposition via Prompt Guidance} 
In this section, we present our method for decomposing complex tasks into simpler sub-tasks, improving confidence, and providing guidance for the DLM during reasoning.

Given an input prompt $P$, instead of asking the model to directly generate an answer, we guide the model to break down the problem into a series of manageable steps. Specifically, we provide the model with an augmented prompt $\hat{P}$ including illustrative two-shot examples showing how a complex question can be divided into a sequence of subtasks. Each subtask is framed with a distinct goal—such as understanding the input, identifying relevant information, or synthesizing the final response—so that guide the model to solve the problem in a structured manner.

At inference time, the model is encouraged to produce its own decomposition for the given input, guided by the example provided in the prompt $\hat{P}$. It then solves the subtasks sequentially, with intermediate outputs serving as context for subsequent steps. In summary, the full algorithm of \modelname is shown in \Cref{alg:td-umcts}. 

\subsection{Theoretical Guarantee}

\begin{table*}[htbp]
\centering
\resizebox{\linewidth}{!}{
\begin{tabular}{l@{\hskip 10pt}l@{\hskip 10pt}l@{\hskip 10pt}l@{\hskip 10pt}l@{\hskip 10pt}l@{\hskip 10pt}l@{\hskip 10pt}c}
\toprule
\textbf{Model} & \textbf{GSM8K} & \textbf{ARC-C} & \textbf{HumanEval} & \textbf{MMLU} & \textbf{DROP} & \textbf{Countdown} & \textbf{Ave. Imprv.} \\
\midrule
LLaDA  & 58.3 & 72.2 & 40.2 & 36.0 & 58.2 & 15.6 &  --- \\

+ Ours & 66.7 \textcolor{darkestgreen}{\scriptsize \shortstack{$\Delta$+8.4\\(14.4\% $\uparrow$)}} & 82.1 \textcolor{darkestgreen}{\scriptsize \shortstack{$\Delta$+9.9\\(13.7\% $\uparrow$)}} & 47.5 \textcolor{darkestgreen}{\scriptsize \shortstack{$\Delta$+7.3\\(18.2\% $\uparrow$)}} & 44.0 \textcolor{darkestgreen}{\scriptsize \shortstack{$\Delta$+7.0\\(19.4\% $\uparrow$)}} & 71.0 \textcolor{darkestgreen}{\scriptsize \shortstack{$\Delta$+12.8\\(22.0\% $\uparrow$)}} & 18.9 \textcolor{darkestgreen}{\scriptsize \shortstack{$\Delta$+3.3\\(21.2\% $\uparrow$)}} & \textcolor{darkestgreen}{\scriptsize \shortstack{$\Delta$+8.1\\(18.2\% $\uparrow$)}}\\

+ Bst5 & 64.2 \scriptsize \shortstack{$\Delta$+5.9\\(10.1\% $\uparrow$)} & 77.2 \scriptsize \shortstack{$\Delta$+5\\(6.9\% $\uparrow$)} & 43.4 \scriptsize \shortstack{$\Delta$+3.2\\(8.0\% $\uparrow$)} & 39.0 \scriptsize \shortstack{$\Delta$+3.0\\(8.3\% $\uparrow$)} & 64.3 \scriptsize \shortstack{$\Delta$+6.1\\(10.4\% $\uparrow$)} & 16.9 \scriptsize \shortstack{$\Delta$+1.3\\(8.3\% $\uparrow$)} & \scriptsize \shortstack{$\Delta$+4.1\\(8.7\% $\uparrow$)} \\

\midrule

LLaDA1.5 & 62.3 & 74.9 & 45.0 & 37.3 & 60.5 & 16.8 & --- \\
+ Ours & 69.0 \textcolor{darkestgreen}{\scriptsize \shortstack{$\Delta$+6.7\\(10.6\% $\uparrow$)}} & 82.8 \textcolor{darkestgreen}{\scriptsize \shortstack{$\Delta$+7.9\\(10.5\% $\uparrow$)}} & 51.0 \textcolor{darkestgreen}{\scriptsize \shortstack{$\Delta$+6.0\\(13.3\% $\uparrow$)}} & 44.5 \textcolor{darkestgreen}{\scriptsize \shortstack{$\Delta$+7.2\\(19.3\% $\uparrow$)}} & 71.1 \textcolor{darkestgreen}{\scriptsize \shortstack{$\Delta$+10.6\\(17.5\% $\uparrow$)}} & 19.2 \textcolor{darkestgreen}{\scriptsize \shortstack{$\Delta$+2.4\\(14.3\% $\uparrow$)}} &\textcolor{darkestgreen}{\scriptsize \shortstack{$\Delta$+6.8\\(14.3\% $\uparrow$)}} \\

+ Bst5 & 64.5 \scriptsize \shortstack{$\Delta$+2.2\\(3.5\% $\uparrow$)} & 77.6 \scriptsize \shortstack{$\Delta$+2.7\\(3.6\% $\uparrow$)} & 48.2 \scriptsize \shortstack{$\Delta$+3.2\\(7.1\% $\uparrow$)} & 39.6 \scriptsize \shortstack{$\Delta$+2.3\\(6.2\% $\uparrow$)} & 65.0 \scriptsize \shortstack{$\Delta$+4.5\\(7.4\% $\uparrow$)} & 17.7 \scriptsize \shortstack{$\Delta$+0.9\\(5.4\% $\uparrow$)} &\scriptsize \shortstack{$\Delta$+2.6\\(5.6\% $\uparrow$)} \\

\midrule
Dream & 60.6 & 78.0 & 40.0 & 35.2 & 55.4 & 13.1 & --- \\
+ Ours & 65.7 \textcolor{darkestgreen}{\scriptsize \shortstack{$\Delta$+5.1\\(8.4\% $\uparrow$)}} & 88.5 \textcolor{darkestgreen}{\scriptsize \shortstack{$\Delta$+10.5\\(13.5\% $\uparrow$)}} & 47.7 \textcolor{darkestgreen}{\scriptsize \shortstack{$\Delta$+7.7\\(19.3\% $\uparrow$)}} & 43.3 \textcolor{darkestgreen}{\scriptsize \shortstack{$\Delta$+8.1\\(23.0\% $\uparrow$)}} & 65.4 \textcolor{darkestgreen}{\scriptsize \shortstack{$\Delta$+10.0\\(18.1\% $\uparrow$)}} & 15.6 \textcolor{darkestgreen}{\scriptsize \shortstack{$\Delta$+2.5\\(19.1\% $\uparrow$)}} 
&\textcolor{darkestgreen}{\scriptsize \shortstack{$\Delta$+7.3\\(16.9\% $\uparrow$)}} \\

+ Bst5 & 62.9 \scriptsize \shortstack{$\Delta$+2.3\\(3.8\% $\uparrow$)} & 80.0 \scriptsize \shortstack{$\Delta$+2.0\\(2.6\% $\uparrow$)} & 43.1 \scriptsize \shortstack{$\Delta$+3.1\\(7.8\% $\uparrow$)} & 38.1 \scriptsize \shortstack{$\Delta$+2.9\\(8.2\% $\uparrow$)} & 57.7 \scriptsize \shortstack{$\Delta$+2.3\\(4.2\% $\uparrow$)} & 15.4 \scriptsize \shortstack{$\Delta$+1.3\\(9.1\% $\uparrow$)} &\scriptsize \shortstack{$\Delta$+2.3\\(6.1\% $\uparrow$)} \\

\midrule
Llama & 59.2 & 70.5 & 45.7 & 44.5 & 60.0 & 3.2 & --- \\
\bottomrule
\end{tabular}
}
\caption{Results on using different backbone models across various benchmarks. For each setting, we report absolute improvements ($\Delta$) and percentage gains relative to the corresponding backbone. The best improvement is highlighted in \textcolor{darkestgreen}{Dark Green}. The average improvement is shown in the last column.}
\label{tb:main}
\end{table*}

In this section, we provide a theoretical guarantee for \modelname. At each decoding step $i$, when revealing a set $z_i$ of tokens independently, the total error decomposes into a model term and a joint-dependence term 
$\mathrm{KL}(q(x_{z_i}\!\mid x_{z<i}) \,\|\, \prod_{\ell\in z_i} q(x_\ell\!\mid x_{z<i}))$.
Following \citet{ben2025accelerated}, this dependence error is upper-bounded by the entropy-gap surrogate
$B(z_i) = \sum_{\ell\in z_i} H(p_\theta(x_\ell\!\mid x_{z<i})) - \max_{\ell\in z_i} H(p_\theta(x_\ell\!\mid x_{z<i}))$.
Summing over $K$ initialization steps yields
$\sum_{i=1}^K \text{DepErr}_i \le \sum_{i=1}^K B(z_i)$.
Our MCTS initialization explicitly minimizes the surrogate cost 
$J(z_{1:K})=\sum_{i=1}^K B(z_i)$ across candidate schedules, 
thus selecting a prefix that achieves the smallest available upper bound on cumulative dependence error among explored options.
This establishes that the proposed initialization is theoretically grounded: although we cannot minimize the true dependence error directly, MCTS optimizes a computable surrogate that provably controls it.
A full derivation is in Appendix~\ref{app_proof}.

\section{Experiments}

In this section, we conduct extensive experiments to evaluate \modelname, guided by the following questions: \textbf{RQ1}: How does our proposed method perform on various datasets compared to state-of-the-art baselines? \textbf{RQ2}: What's the contribution of each component in our framework to the overall performance? \textbf{RQ3}: How does the model's performance vary with different hyperparameter settings?

\subsection{Experimental Setup}
We conduct experiments on six widely-used datasets: GSM8K \cite{cobbe2021training}, ARC-C \cite{clark2018think}, HumanEval \cite{chen2021evaluating}, MMLU \cite{hendrycks2020measuring}, DROP \cite{dua2019drop} and Countdown \cite{tinyzero}. Experiments are conducted on models with a similar scale (7B-8B parameters) to ensure a fair comparison.
We evaluate our method on three backbone DLMs, LLaDA-8B-Instruct \cite{nie2025large}, LLaDA1.5-8B \cite{zhu2025llada}, and Dream-7B \cite{ye2025dream}, and compare their performance against (i) the original models, (ii) the original models with a Best-of-5 decoding strategy (generating five samples and selecting the majority answer, bst5), and (iii) the original models with our method. We additionally report results for Llama-3 8B \cite{touvron2023llama} as an autoregressive LLM baseline. We use accuracy for GSM8K, ARC-C, and MMLU; Exact Match for Countdown; pass@1 for HumanEval; and F1 for DROP. For MCTS initialization on all datasets, we set the MCTS initialization candidate length $L_c$ to $20$, $K_1=3$, $K_2=5$, and the stability constant $\varepsilon=10^{-8}$. The generation length for DLMs is $256$. For task-decomposition prompting, we set the number of subtasks to $3$. We set the candidate size $C=3$. All the experiments are conducted on 2 NVIDIA A100 GPUs with 40GB memory, and the random seed is set to $1$. Detailed settings are provided in Appendix~\ref{app:exp_setting}, and computational costs in Appendix~\ref{app:additional_exp}.

\begin{table}[htbp]

\resizebox{0.5\textwidth}{!}{
\centering
    \begin{tabular}{llll}
\toprule
\textbf{Method}                               & \textbf{ARC-C} & \textbf{HumanEval} & \textbf{DROP}  \\
\midrule
\textbf{Ours}                        & 82.1 & 47.5     & 71.0 \\
\textbf{W/o MCTS}                            & 81.0  & 44.3     & 64.9 \\
\textbf{W/o T. Dcp.}                         & 77.3 & 43.4      & 65.7 \\
\textbf{W/o Cf.+t2} & 75.0  & 40.9     & 60.1 \\
\textbf{Llada}                               & 72.2 & 40.2     & 58.2\\
\bottomrule
\end{tabular}
}

\caption{Ablation study results on ARC-C, HumanEval, and DROP using LLaDA as the backbone model.}
\label{tb:ablation}
\end{table}

\subsection{Effectiveness of MEDAL}

We evaluate our method on the five datasets using three different backbone DLMs. The results are shown in Table~\ref{tb:main}. We observe that our method consistently improves the performance of all backbone models across all datasets, with up to  
 $18.2$\% average and $8.1$ absolute improvement, indicating the effectiveness and generality of our approach. Notably, even a model that underperforms Llama (i.e., Llada, which lags behind Llama on 4 out of 5 baselines in Table~\ref{tb:main}) achieves comparable or superior results on most datasets when equipped with our method. These results highlight the potential of DLMs when guided with appropriate strategies. The standard deviation is reported in Table~\ref{app_tb:std}.

\subsection{Ablation Study}

To understand the contribution of each component in our framework, we conduct ablation studies on ARC-C, HumanEval and DROP on backbone model LLaDA. To evaluate the impact of MCTS, we remove it from our framework and let the model generate answers directly based on the decomposed tasks (\textbf{W/o MCTS}). To assess the importance of task decomposition, we eliminate this step and have the model generate answers directly from the original prompt using MCTS (\textbf{W/o T. Dcp.}). Additionally, we remove the confidence-adjusted score and using only use top-2 margin (\textbf{W/o Cf.+t2}) to select the candidate tokens during MCTS.

The results are presented in Table \ref{tb:ablation}. Overall, removing any component leads to a performance drop arcoss all datasets. Specifically, removing or replacing the confidence-adjusted score results in the largest performance drop, highlighting its critical role in selecting tokens to unmask. Moreover, eliminating MCTS also causes a performance drop, demonstrating its effectiveness in exploring multiple generation paths and selecting the most promising ones during initialization. 

\subsection{Hyperparameter Analysis}

\begin{table}[htbp]
\resizebox{0.5\textwidth}{!}{
\centering
\begin{tabular}{llll}
\toprule

\textbf{Tasks No.}  & \textbf{ARC-C} & \textbf{HumanEval} & \textbf{DROP}  \\
\midrule
\textbf{1}                                    & 79.8 & 44.0      & 69.2 \\
\textbf{3}             & 82.1 & 47.5     & 71.0 \\
\textbf{5}                                    & 76.5 & 42.9     & 62.8 \\
\textbf{10}                                   & 75.5 & 41.9     & 58.0 \\
\bottomrule
\end{tabular}
}
\caption{Performance with different numbers of decomposed tasks on ARC-C, HumanEval, and DROP using LLaDA as the backbone model.}

\label{tb:hyper_task}
\end{table}

In this section, we analyze the impact of key hyperparameters in our framework, including the number of decomposed tasks, the number of candidate tokens selected $K_2$ during MCTS, and MCTS initialization candidate length $L_c$. 
We vary the number of decomposed tasks and evaluate the performance on ARC-C, HumanEval, and DROP using LLaDA as the backbone model. The results are shown in Table \ref{tb:hyper_task}. We observe that decomposing the task into 3 subtasks yields the best performance. Decomposing into too few (1) or too many (5 or 10) subtasks leads to a performance drop, indicating that an optimal level of decomposition is crucial for balancing complexity and guidance. 

Then we vary the number of candidate tokens selected during MCTS and evaluate the performance on the same datasets. The results are shown in Table \ref{tb:hyper_cand}. We find that selecting 5 candidate tokens during MCTS achieves the best performance across all datasets.
Selecting too few (1 or 3) candidate tokens would made the exploration space too small, limiting the potential of finding better generation paths. On the other hand, selecting too many (10) tokens would introduce too much noise, making it harder for the model to focus on the most promising paths. Therefore, we find that selecting 5 candidate tokens works well which is essential for effective exploration and exploitation during MCTS. 

\begin{figure*}[t]
    \centering
    \includegraphics[width=\textwidth]{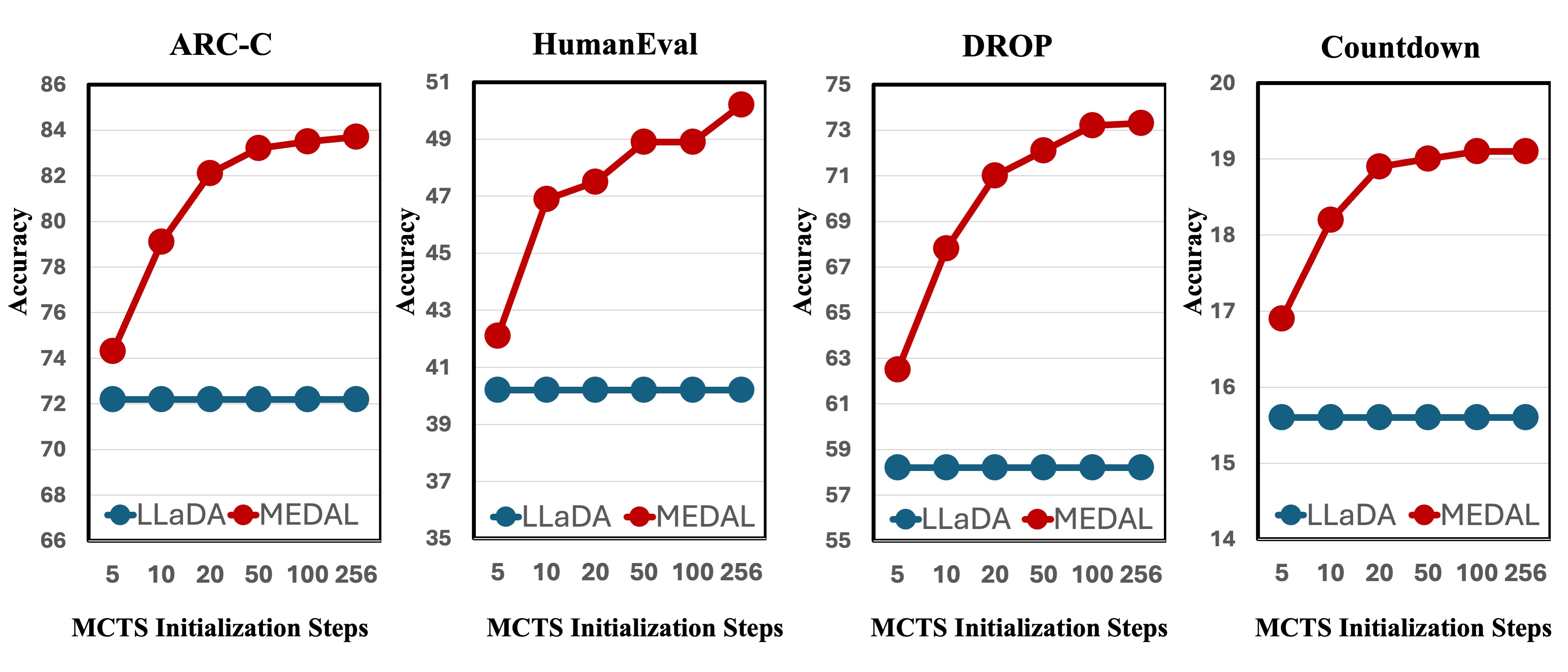}

    \caption{Performance of \modelname (in red) with varying MCTS init length $L_c$ on ARC-C, HumanEval, DROP and Countdown using LLaDA as the backbone model. The blue line denotes the performance of LLaDA without our method. The x-axis indicates the number of MCTS initialization steps, and the y-axis shows the accuracy.}
    \label{fig:mcts_step}
\end{figure*}

Finally, we vary the number of MCTS initialization candidate length $L_c$ and evaluate the performance on data, including ARC-C, HumanEval, DROP and Countdown, for generating sentences of length $256$. The results are shown in Figure \ref{fig:mcts_step}. We observe that increasing $L_c$ can improve the performance, and the performance gain tends to saturate after 20 steps. It indicates that using MCTS for initialization is sufficient, and it is a feasible approach for effective generation. This result suggests that the most critical generation trajectory decisions occur early, meaning that extended search beyond this point yields diminishing returns relative to the computational cost.

\begin{table}[]
    \centering
\resizebox{0.5\textwidth}{!}{
\begin{tabular}{llll}
\toprule

\textbf{MCTS Cand.} & \textbf{ARC-C} & \textbf{HumanEval} & \textbf{DROP}  \\
\midrule

\textbf{1}                         & 79.2 & 44.4     & 65.5 \\
\textbf{3}                         & 81.5 & 46.4     & 70.1 \\
\textbf{5}                         & 82.1 & 47.5     & 71.0 \\
\textbf{10}                        & 82.9 & 46.6     & 70.5\\
\bottomrule
\end{tabular}
}
\caption{Performance with different numbers of candidate tokens $K_2$ during MCTS on ARC-C, HumanEval, and DROP using LLaDA as the backbone model.}

\label{tb:hyper_cand}
\end{table}

\subsubsection{Case Study: Agentic Workflows}
In this section, we present a case study exploring the potential of using DLMs in an agentic setting. We integrate our method with the ADAS~\cite{hu2024automated}, which utilizes an LLM to automatically invent building blocks and design powerful agentic systems to solve given tasks. We replace the LLMs in ADAS with our DLM (LLaDA with our method) and compare the performance with the original ADAS using LLaDA and Llama as backbones. The results on DROP and MMLU are shown in Table \ref{tb:agentic}. We observe that integrating our method with ADAS leads to further performance improvement compared to using LLaDA and LLama, demonstrating our method's ability to enhance the reasoning and planning capabilities of DLMs in complex agentic settings. These findings suggest that DLMs, when equipped with effective strategies like ours, can be powerful tools for building intelligent agents capable of solving challenging tasks.

\begin{table}[]
    \centering
\resizebox{0.38\textwidth}{!}{

\begin{tabular}{lll}
\toprule

\textbf{Method}       & \textbf{DROP}  & \textbf{MMLU}  \\
\midrule

\textbf{Ours + ADAS}  & 73.0 & 46.5 \\
\textbf{LLada + ADAS} & 71.2 & 41.0  \\
\textbf{Llama + ADAS} & 65.2 & 45.2 \\
\textbf{LLada} & 58.2 & 36.0  \\

\textbf{Llama}        & 60.0   & 44.5 \\
\bottomrule
\end{tabular}
}

\caption{Results on DROP and MMLU using different methods in an agentic setting.}

\label{tb:agentic}
\end{table}

\section{Related works}

\paragraph{Diffusion Language Models.}
Diffusion models are a powerful class of generative models \cite{podell2023sdxl, rombach2022high, li2025survey, zhang2023adding, li2025diffusion}. Building on their success, DLMs have emerged as promising alternatives for text generation. One line of work adapts continuous diffusion to discrete text via continuous relaxations \cite{li2022diffusion, strudel2022self}, while another operates directly in the discrete token space, corrupting text by masking or token replacement \cite{he2022diffusionbert, austin2021structured}. Leveraging mature scaling techniques, large-scale DLMs have been developed \cite{nie2025large, gong2024scaling, nie2024scaling}, achieving performance competitive with AR models \cite{touvron2023llama}. Despite strong generative ability, DLMs often struggle with controllability and reasoning in deciding the token order to unmask tokens during inference. To address this, we propose a principled search-based framework with MCTS to enhance DLM generation capabilities.

\paragraph{Enhancing Reasoning Capabilities of Diffusion Language Models.}
To improve the reasoning performance of DLMs \citet{kim2025train} propose incorporating model uncertainty into the diffusion process to enhance generation quality. Entropy-based planning methods have been introduced \cite{ben2025accelerated, ye2025dream} to better capture the model’s confidence. Planner-guided generation is studied in \cite{peng2025path}, enabling token-level refinement during sampling. More advanced reinforcement learning (RL) approaches further boost performance \cite{zekri2025fine, zhu2025llada, gong2025diffucoder}. Particularly, Diffu-GRPO \cite{zhao2025d1} applies policy-gradient RL to DLMs via a mean-field approximation and prompt masking, while TraDo \cite{wang2025revolutionizing} aligns DLM inference trajectories with training objectives by optimizing the sampling process. Additionally, wd1 \cite{tang2025wd1} reformulates the RL objective as a weighted likelihood to avoid biased policy ratios, improving reasoning accuracy. While effective, these methods rely on additional training that involves substantial computational cost and engineering overhead. In contrast, we enhance DLM reasoning purely at inference time, making our approach more efficient and practical for real-world applications.

\section{Conclusion}
In this work, we introduced \modelname, a framework that casts diffusion language model inference as a structured search problem and integrates it with MCTS. By applying MCTS during initialization, \modelname\ explores promising unmasking trajectories before refinement. Its confidence-guided filtering and information-gain reward enable efficient, targeted search and improve global certainty. Across multiple benchmarks, \modelname\ consistently outperforms existing inference strategies, achieving gains of up to 22.0\%.

\section{Limitation}
While our study demonstrates the effectiveness of confidence-guided MCTS initialization for diffusion language models, several limitations remain. First, we primarily evaluate unimodal text-based DLMs; extending our framework to multimodal DLMs (e.g., vision-language or audio-language models) would provide a more comprehensive assessment of its generality. Second, our current experiments focus on standalone inference; integrating the method into agentic settings, where reasoning and decision-making unfold over multiple steps and interactions, poses both challenges and opportunities for future exploration.

\section{Potential Risks}
This work does not introduce additional ethical or societal risks beyond those associated with existing large language models. While our method utilizes test-time scaling to enhance generation quality, it operates by allocating compute to resolve uncertainty within the fixed model distribution. Consequently, it does not alter the model's fundamental objectives, training data, or safety alignment.

\section{Acknowledgements}
We used generative AI tools to assist in language polishing and writing refinement. All conceptual, methodological, and experimental contributions are solely those of the authors.

\bibliographystyle{acl_natbib}
\bibliography{custom}


\clearpage

\appendix

\section{Appendix}
\label{sec:appendix}

\subsection{Experiment Setting}
\label{app:exp_setting}

In this section, we provide more details about the experimental settings. We use three different backbone models: LLaDA-8B-Instruct \cite{nie2025large}, LLaDA1.5-8B \cite{zhu2025llada}, and Dream-7B \cite{ye2025dream}, and compare the results with the original models without our method. We also include Llama3-8B \cite{touvron2023llama} as an AR LLM baseline for comparison. The details of the models are as follows:
\begin{itemize}
    \item \textbf{LLaDA-8B-Instruct.} LLaDA is a large-scale diffusion language model (DLM) that replaces the traditional autoregressive next-token prediction paradigm with a masked diffusion process. Instead of generating text sequentially from left to right, LLaDA learns to reconstruct corrupted sequences by iteratively predicting masked tokens. The model defines a forward process that randomly masks tokens and a reverse process—parameterized by a transformer without causal masking—that predicts the original tokens, allowing bidirectional context modeling. Trained under a principled likelihood bound, LLaDA achieves scalable probabilistic inference comparable to large autoregressive models such as Llama3. It supports parallel token refinement, enabling faster generation and better global coherence. After pre-training on 2.3T tokens and fine-tuning on 4.5M instruction pairs, LLaDA demonstrates strong performance across reasoning, coding, and multilingual tasks, rivaling or surpassing autoregressive LLMs of similar scale while also addressing their limitations in efficiency, reversibility, and bidirectional reasoning.
    \item \textbf{LLaDA1.5-8B.} LLaDA1.5 is an advanced version of LLaDA, specifically designed to be better aligned with human preferences through reinforcement learning. The key innovation is a framework called Variance-Reduced Preference Optimization (VRPO), which addresses the primary challenge of applying direct preference optimization to diffusion models: the high variance in their likelihood estimations. VRPO introduces theoretically-grounded, unbiased techniques, such as optimal Monte Carlo budget allocation and antithetic sampling, to significantly reduce this variance, leading to more stable and effective training. 
    \item \textbf{Dream-7B.} Dream-7B is a diffusion large language model that generates text by iteratively refining sequences in parallel rather than sequentially like autoregressive models. It is trained using two key techniques: initializing its weights from a pre-trained AR model and employing a context-adaptive token-level noise rescheduling mechanism. While achieving performance competitive with leading AR models like Qwen2.5 7B on general, mathematical, and coding benchmarks, Dream 7B demonstrates substantial abilities on complex planning tasks such as Sudoku and trip planning.
\end{itemize}

The data we used for evaluation includes six widely used datasets:
GSM8K \cite{cobbe2021training}, ARC-C \cite{clark2018think}, HumanEval \cite{chen2021evaluating}, MMLU \cite{hendrycks2020measuring}, DROP \cite{dua2019drop} and Countdown \cite{tinyzero}. The details of the datasets are as follows:
\begin{itemize}
  \item \textbf{GSM8K.} A dataset to evaluate the mathematical and scientific reasoning capabilities of large language models. The dataset consists of high-quality, grade-school-level math word problems that require multiple steps to solve. These problems are designed to test a model's ability to perform 

  \item \textbf{ARC-C.} The ARC Challenge (ARC-C) is a subset of the AI2 Reasoning Challenge (ARC) dataset, which is designed to evaluate the reasoning capabilities of AI systems. The ARC-C consists of multiple-choice questions, requiring deeper reasoning and understanding of scientific concepts. The questions cover a wide range of topics in science and are intended to test a model's ability to apply knowledge rather than just recall facts.

  \item \textbf{HumanEval.} A benchmark dataset for evaluating the programming capabilities of models by presenting them with a series of coding challenges. These challenges require the model to generate functional code from natural language descriptions, often in a zero-shot setting where no examples are provided.

  \item \textbf{MMLU.} The Massive Multitask Language Understanding (MMLU) benchmark is designed to measure the general knowledge, reasoning, and commonsense abilities of large language models. It consists of multiple-choice questions, including humanities, social sciences, STEM fields, and more. The benchmark tests a model's ability to understand and generate text in various contexts, requiring both factual knowledge and reasoning skills.

  \item \textbf{DROP.} A reading comprehension benchmark that requires discrete reasoning over paragraphs of text. The dataset consists of questions that necessitate operations such as addition, counting, and sorting to arrive at the correct answer. 

  \item \textbf{Countdown.} A benchmark that contains arithmetic reasoning problems where the model is given three or four numbers and must construct an expression that uses each number exactly once with basic operations (+, –, ×, ÷) to reach a target integer. Each example includes a list of numbers and a target.
  
\end{itemize}

\begin{table*}[t]
\centering
\begin{tabular}{l@{\hskip 10pt}l@{\hskip 10pt}l@{\hskip 10pt}l@{\hskip 10pt}l@{\hskip 10pt}l@{\hskip 10pt}c}
\toprule
\textbf{Model} & \textbf{GSM8K} & \textbf{ARC-C} & \textbf{HumanEval} & \textbf{MMLU} & \textbf{DROP}  \\
\midrule
LLaDA  & $\pm 1.36$ & $\pm 1.48$ & $\pm 2.51$ & $\pm4.34$ & $\pm3.17$ \\

+ ours & $\pm 2.37$  & $\pm 2.03$  & $\pm 2.98$  & $\pm 3.67$ & $\pm 3.25$  \\
\midrule
LLaDA1.5 & $\pm 3.05$ & $\pm 2.00$ & $\pm 3.22$ & $\pm 4.00$ & $\pm 2.12$ &  \\
+ ours & $\pm 3.27$ & $\pm 3.65$ & $\pm 2.44$& $\pm 3.77$ & $\pm 4.44$ &  \\
\midrule
Dream & $\pm 3.32$ & $\pm 2.09$ & $\pm 3.04$ & $\pm 2.27$ & $\pm 3.38$  \\
+ ours & $\pm 2.93$  & $\pm 3.32$  & $\pm 3.85$  & $\pm 2.41$ & $\pm 4.00$  \\
\midrule
Llama & $\pm 3.05$ & $\pm 3.31$ & $\pm 4.22$ & $\pm 3.02$ & $\pm 2.03$  \\
\bottomrule
\end{tabular}
\caption{Results on standard deviation using different backbone models across various benchmarks.}
\label{app_tb:std}
\end{table*}

\begin{figure*}[!h]
    \centering
    \includegraphics[width=\textwidth]{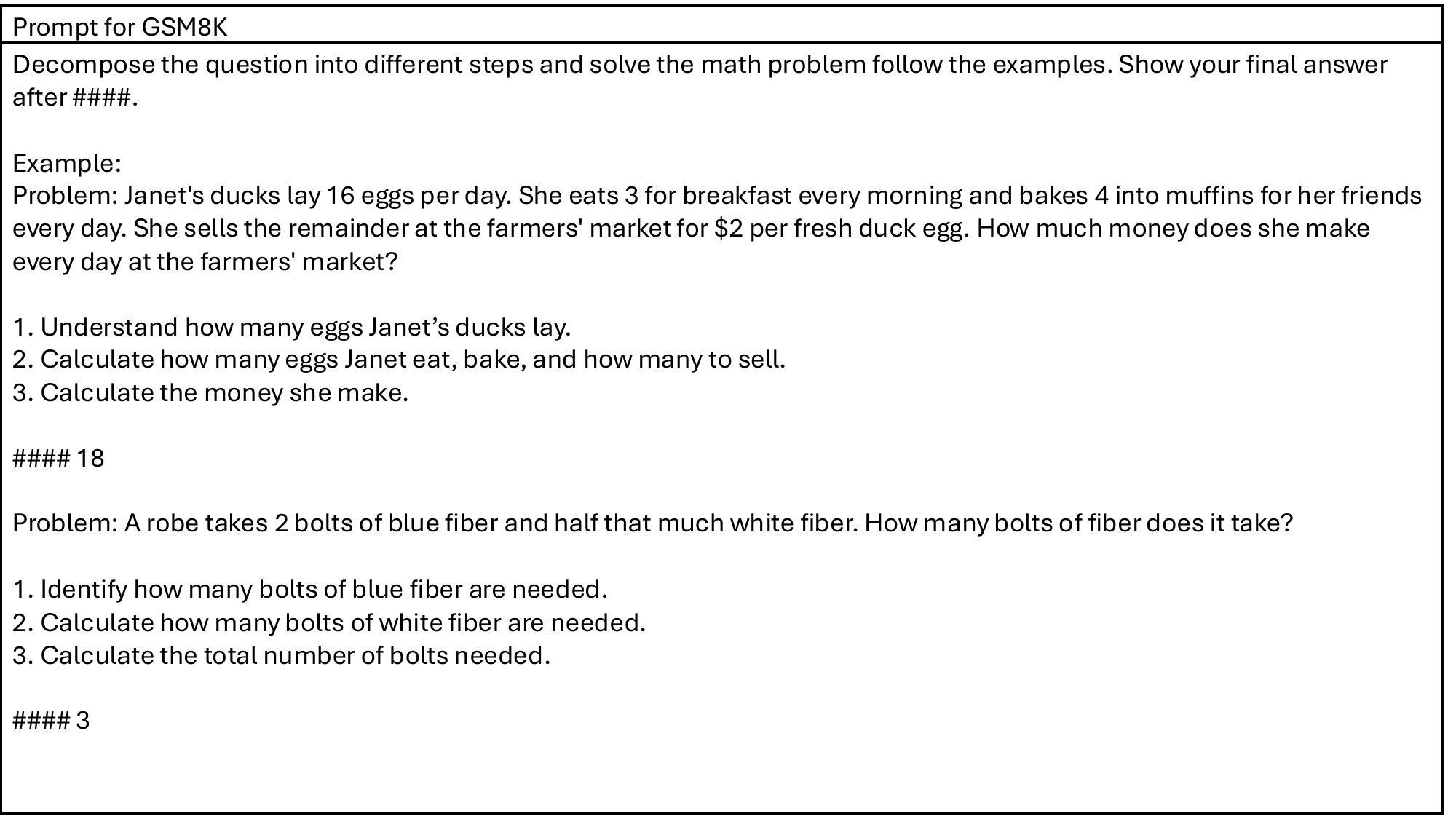}
    \caption{Prompt used for GSM8K}
    \label{fig:propt_gsm8k}
\end{figure*}

\begin{figure*}[]
    \centering
    \includegraphics[width=\textwidth]{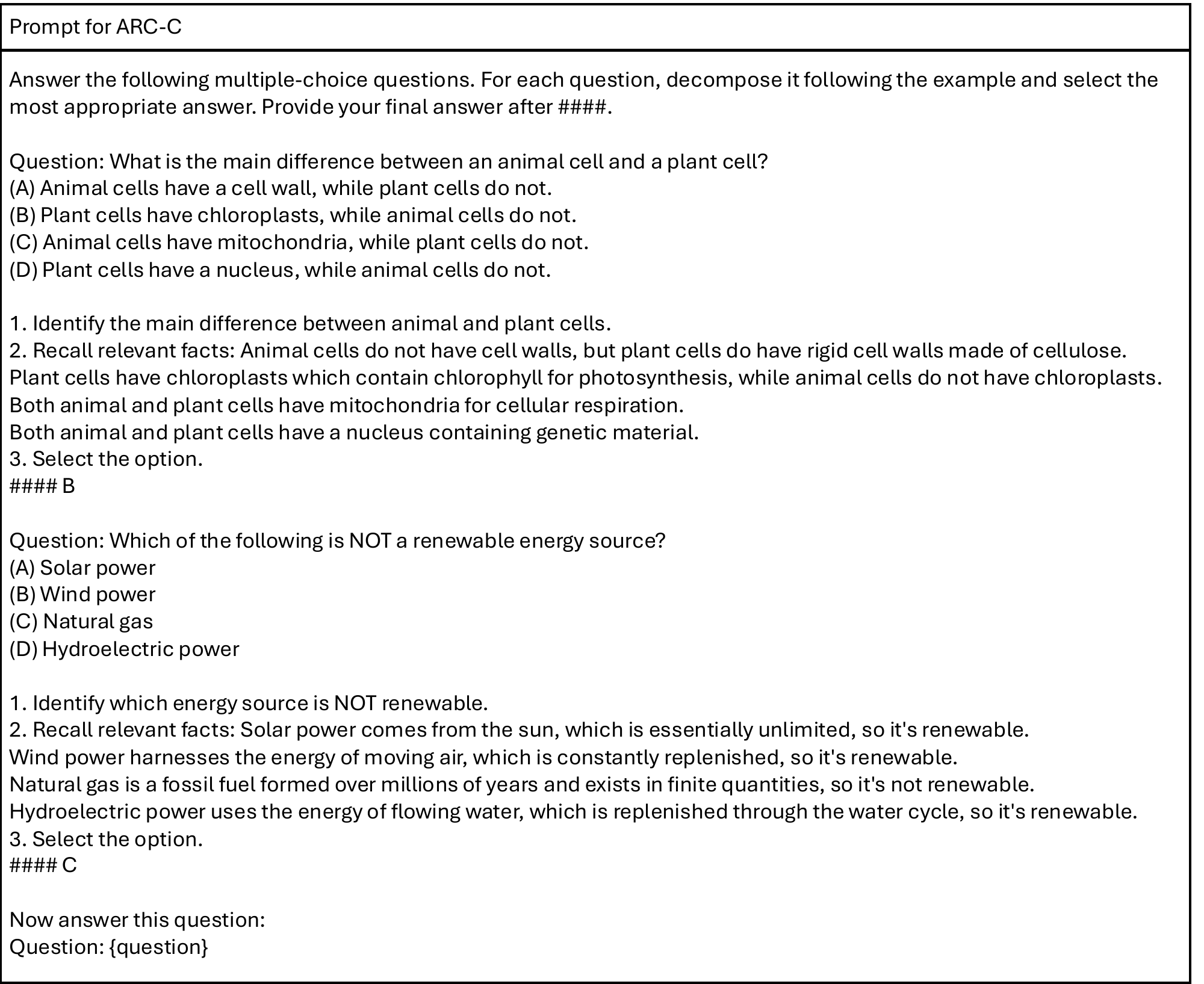}
    \caption{Prompt used for ARC-C}
    \label{fig:propt_arc_c}
\end{figure*}

\begin{figure*}[]
    \centering
    \includegraphics[width=\textwidth]{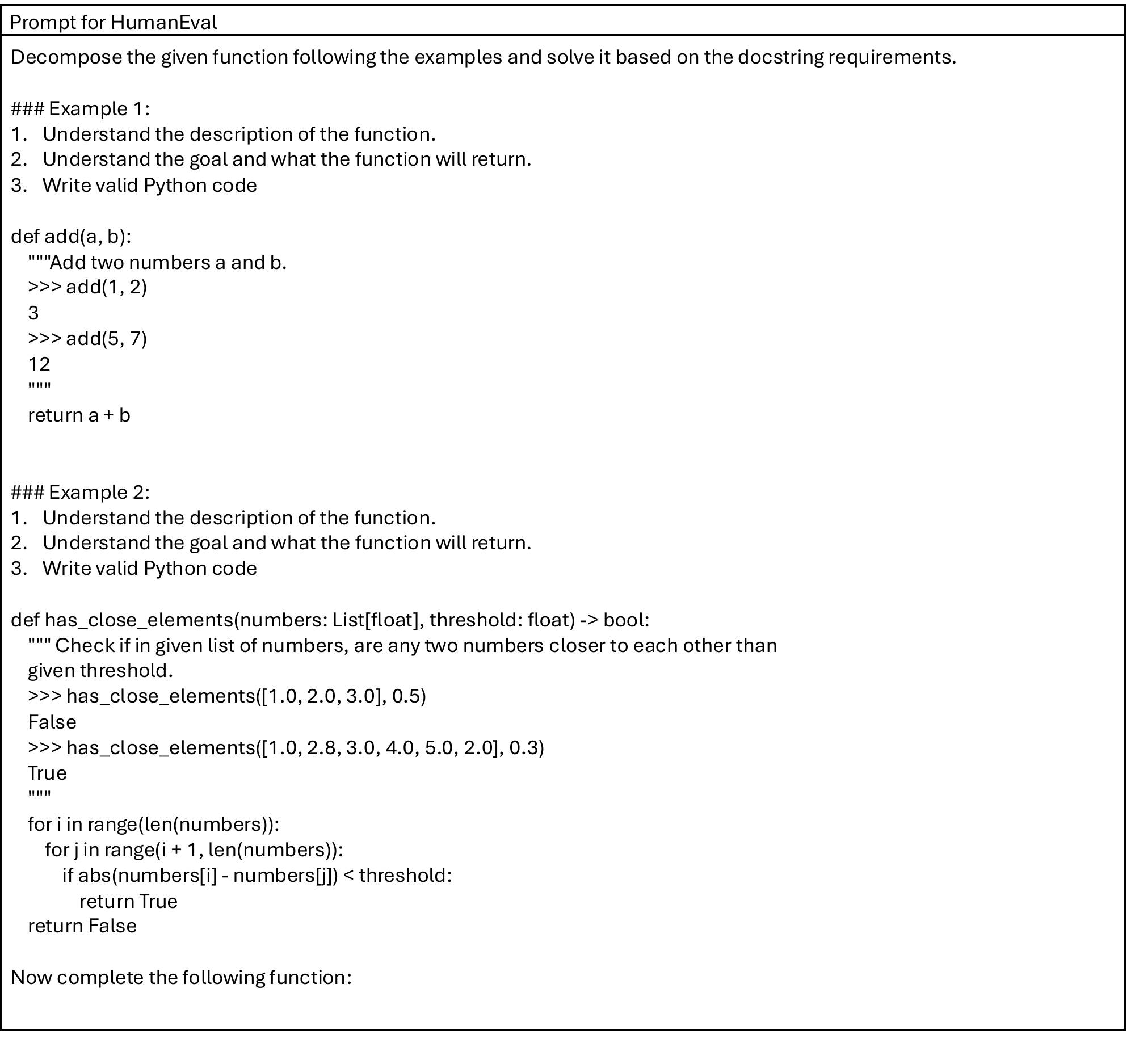}
    \caption{Prompt used for HumanEval}
    \label{fig:propt_human}
\end{figure*}

\begin{figure*}[]
    \centering
    \includegraphics[width=\textwidth]{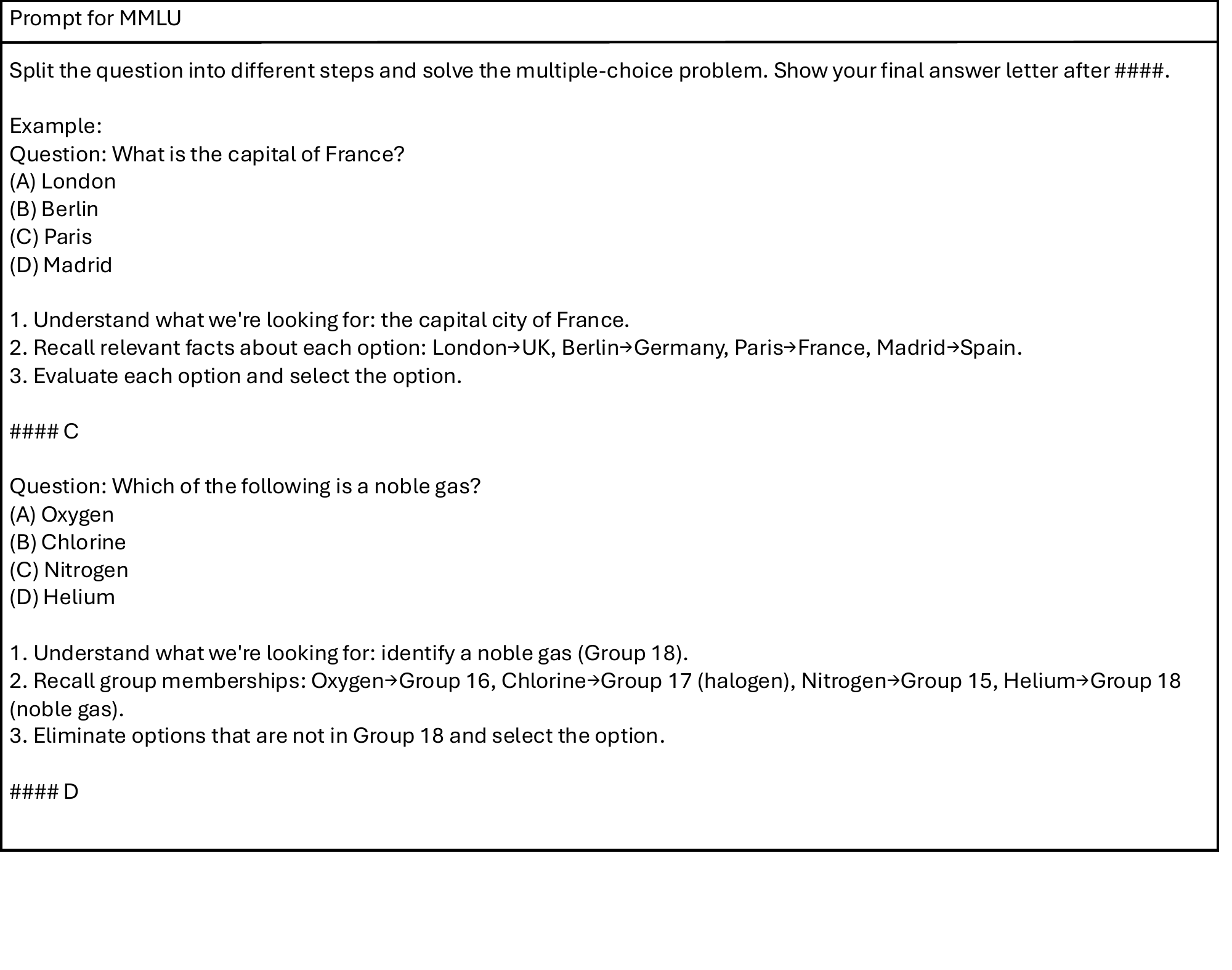}
    \caption{Prompt used for MMLU}
    \label{fig:propt_mm}
\end{figure*}

\begin{figure*}[]
    \centering
    \includegraphics[width=\textwidth]{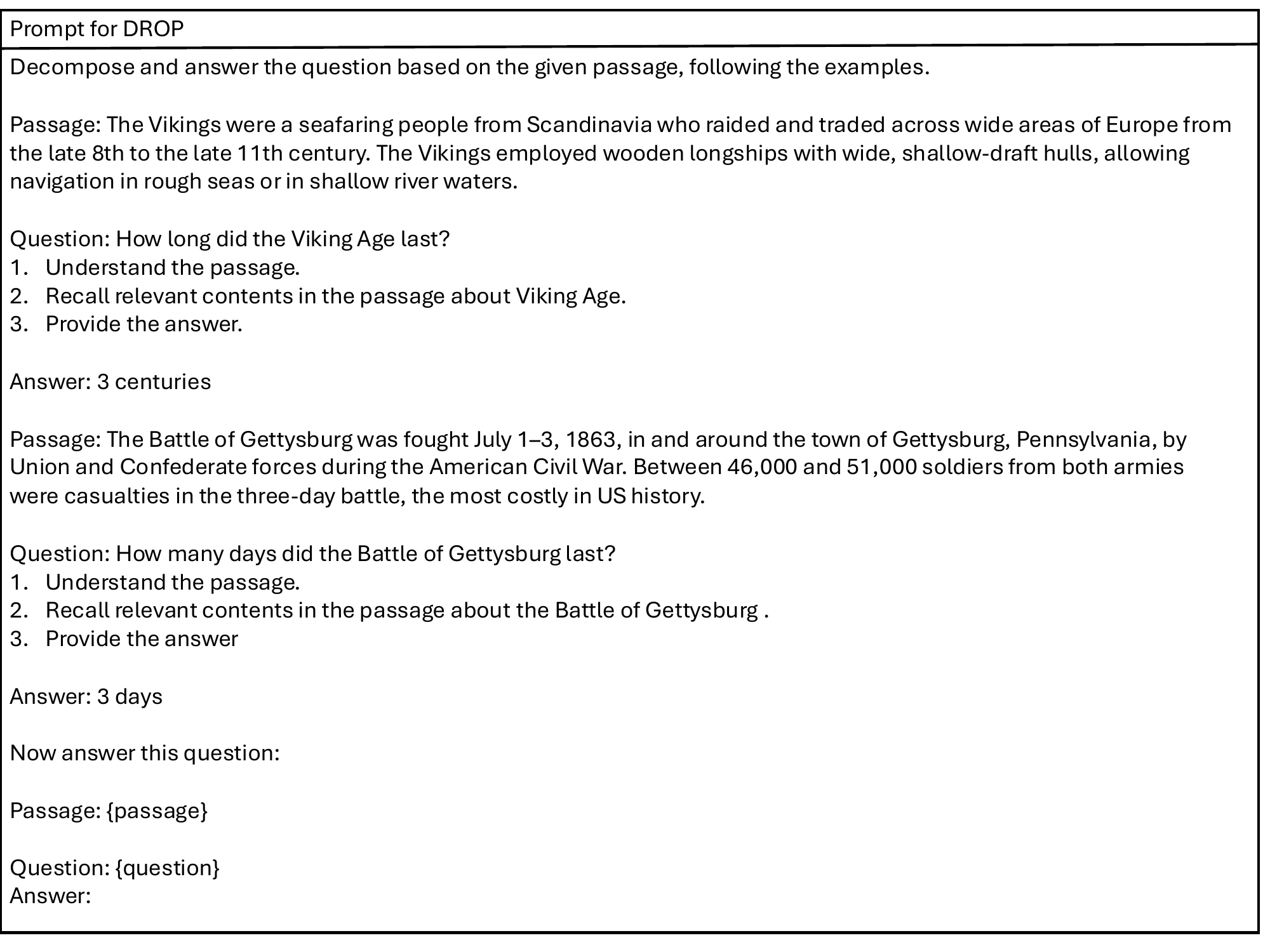}
    \caption{Prompt used for DROP}
    \label{fig:propt_drop}
\end{figure*}

The prompt we used for each dataset is shown in~\Cref{fig:propt_gsm8k}, \ref{fig:propt_arc_c}, \ref{fig:propt_human} , \ref{fig:propt_mm}, and \ref{fig:propt_drop}.

\begin{figure*}[t]
    \centering
    \includegraphics[width=\textwidth]{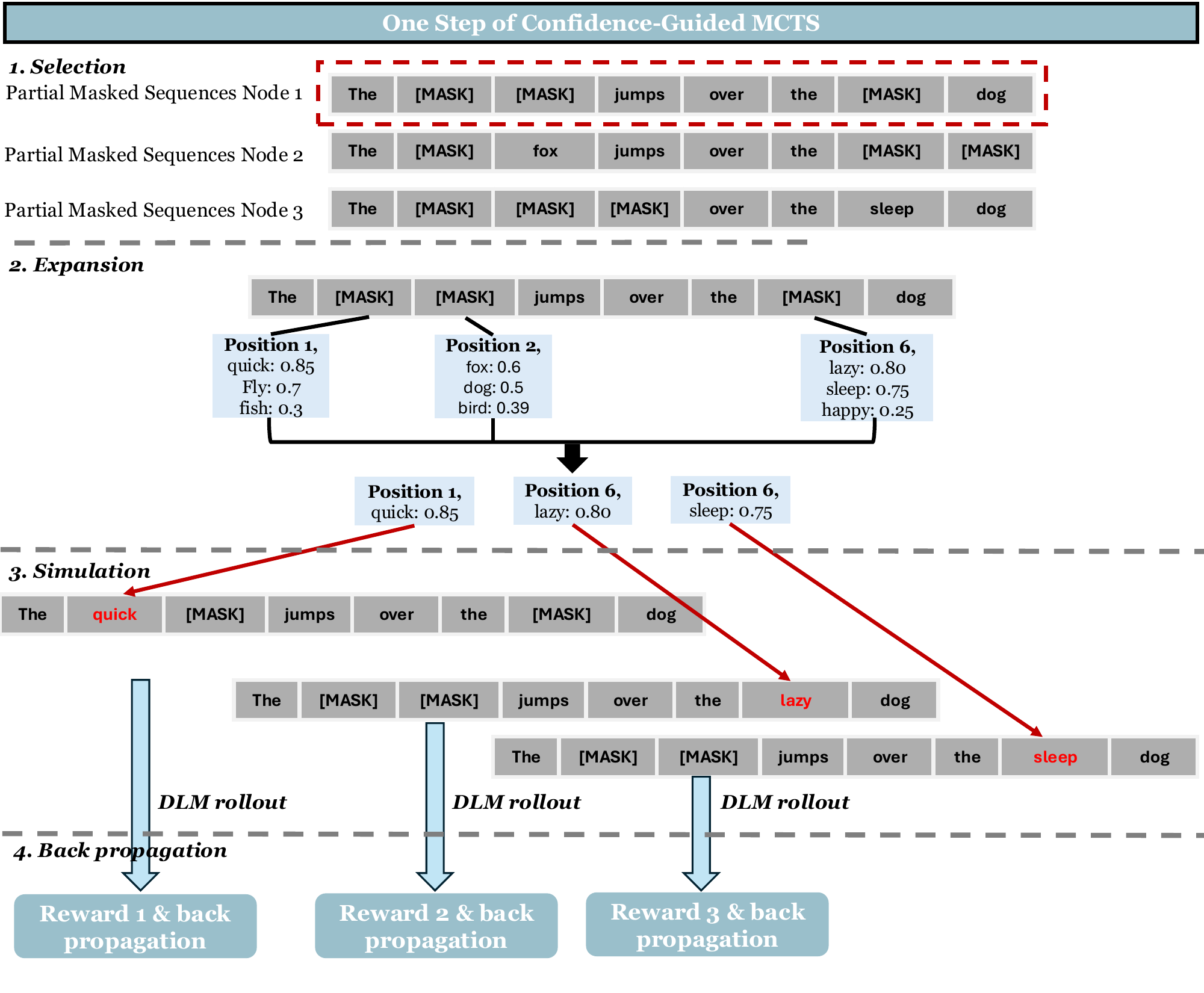}
    \caption{A single-step illustrative example of applying MCTS to DLM inference}
    \label{fig:mcts_step_eg}
\end{figure*}

\subsection{Additional Experiments}
\label{app:additional_exp}

In this section, we report the standard deviation and the computing overhead.

\textbf{Standard Deviation. }
The standard deviation of all the methods across various datasets. The results are shown in Table~\ref{app_tb:std}. Overall, the results show that incorporating our method generally stabilizes model performance by reducing variance across benchmarks, particularly on reasoning-intensive tasks such as HumanEval, MMLU, and DROP. 

\textbf{Computational Costs. } We evaluate computational overhead by comparing a standard LLaDA run with LLaDA using Best-of-15 (Bst15) decoding on GSM8K. The results are reported in Table \ref{table:computing_cost}. As shown, MEDAL achieves comparable runtime to Bst15 while delivering better accuracy, indicating that MEDAL allocates inference-time compute more effectively under constrained resources.

\begin{table}[t]
\centering
\begin{tabular}{lll}
\toprule
      & Running time & Accuracy \\
      \midrule
LLaDA & c=9.64s      & 58.3  \\
+bst15 & 15c          & 65.3  \\
+ours & 12.3c        & 66.7 \\ 
\bottomrule
\end{tabular}
\caption{Computational costs and accuracy on GSM8K using different backbone models, with accuracy reported in the leftmost column.}
\label{table:computing_cost}
\end{table}

\subsection{MCTS for DLMs Inference Example}
\label{app_mcts_example}
We provide a single-step illustrative example of applying MCTS to DLM inference in Figure~\ref{fig:mcts_step_eg}. (1) Selection: among three partially unmasked sequence nodes, we pick the one with the highest UCB score (Node~1). (2) Expansion: given the model logits at Node~1, we compute confidence-adjusted scores for all masked positions and construct the candidate action set $\mathcal{A}_t'$. For each candidate $a\in\mathcal{A}_t'$ (e.g., inserting the token ``quick'' at position 1), we apply the edit and create a new child node. (3) Simulation: from each expanded node, we unmask the remaining tokens using the DLM to obtain a completed sequence. (4) Backpropagation: we compute the information-gain reward (\Cref{info_gain}) for each simulation outcome and propagate it up the tree, updating value estimates for all parent nodes along the path to the root.

\subsection{Formal Derivation of the MCTS Initialization Guarantee}
\label{app_proof}
\paragraph{Setup and notation.}
Let $\mathbf{x}=(x_1,\ldots,x_n)$ be a sequence of discrete tokens.
We denote by $q$ the (unknown) true data distribution and by $p_\theta$ the model distribution.
Decoding proceeds in steps $i=1,2,\ldots$; at step $i$ we reveal a (possibly multi-token) index set $z_i\subseteq\{1,\dots,n\}$,
and write $x_{z<i}$ for all tokens revealed before step $i$.
Let $U$ denote a set of indices of masked tokens.
For such a set $U$ and context $x_{z<i}$, we define the model conditional entropy as
$H_\theta(x_\ell\!\mid\!x_{z<i})\!:=\!H(p_\theta(x_\ell\!\mid\!x_{z<i}))$, and the entropy-gap surrogate

\begin{align}
B(U\mid x_{z<i})
&:= \sum_{\ell\in U} H_\theta(x_\ell\mid x_{z<i}) \nonumber\\
&- \max_{\ell\in U} H_\theta(x_\ell\mid x_{z<i}). 
\end{align}
The term $B(U\mid x_{z<i})$ measures the total uncertainty of the tokens in $U$, penalized by subtracting the largest single-token entropy. Intuitively, it is small when one token in $U$ dominates the uncertainty and large when multiple tokens are simultaneously high-entropy, making it a useful surrogate for the dependence error incurred by unmasking them together. To simplify the notation, we write $B(z_i):=B(z_i\mid x_{z<i})$ when the context is clear.

\paragraph{Per-step decomposition}
When sampling $x_{z_i}$ independently given $x_{z<i}$, the per-step KL divergence decomposes into a token-wise model error and a joint-dependence error\cite{ben2025accelerated}:

\begin{align}
&\underbrace{\sum_{\ell\in z_i}
   \mathrm{KL}\!\big(q(x_\ell\mid x_{z<i}) \,\|\, 
   p_\theta(x_\ell\mid x_{z<i})\big)}_{\text{model error}}
\nonumber\\
&\quad+\;
\underbrace{\mathrm{KL}\!\Big(q(x_{z_i}\mid x_{z<i}) \,\Big\|\,
   \textstyle\prod_{\ell\in z_i} q(x_\ell\mid x_{z<i})\Big)}
   _{\text{dependence error }:=\,\DepErr_i},
\end{align}
where the first term, model error, sums the KL divergence for each $l \in z_i$, quantifying the per-token discrepancy between the model’s conditional distribution and the true conditional given the current context, and
the joint-dependence error quantifies the penalty from sampling all tokens in $z_i$ independently, measuring how far the true joint conditional distribution $q(x_{z_i}\mid x_{z<i})$ is from the product of its marginals. It captures the correlations among tokens in $z_i$ that are ignored when they are sampled independently, rather than jointly conditioned on one another.
Our analysis focuses on the dependence term $\DepErr_i$.

\paragraph{Standing assumption.}
\begin{assumption}[Entropy-gap upper bound]\label{asmp:gap}
For every step $i$ and context $x_{z<i}$,
\begin{align}
\DepErr_i
&= \mathrm{KL}\!\Big(q(x_{z_i}\mid x_{z<i}) \,\Big\|\,
      \textstyle\prod_{\ell\in z_i} q(x_\ell\mid x_{z<i})\Big) \nonumber\\
&\le B(z_i).
\end{align}

\end{assumption}
\noindent
Our contribution is to show how the MCTS initializer is designed to minimize the RHS (a computable surrogate), thereby controlling the cumulative dependence error.

\paragraph{Cumulative bound.}
\begin{lemma}[Prefix dependence is bounded by cumulative entropy gaps]\label{lem:cum}
For any $K\ge 1$,
\[
\sum_{i=1}^K \DepErr_i \;\le\; \sum_{i=1}^K B(z_i).
\]
\end{lemma}
\begin{proof}
Apply Assumption~\ref{asmp:gap} to each step and sum over $i=1,\dots,K$.
\end{proof}

\paragraph{Search space and surrogate objective.}
Fix $K\!\ge\!1$ and let $\mathcal{S}_K$ be the set of feasible $K$-step schedules $z_{1:K}=(z_1,\ldots,z_K)$
(e.g., obeying any architectural or mask constraints).
Define the \emph{surrogate cost} of a schedule by
\[
J(z_{1:K}) \;:=\; \sum_{i=1}^K B(z_i).
\]
(Optionally, one may add a tokenwise uncertainty term; see Remark~\ref{rem:weights} below.)

\paragraph{MCTS estimator and selection rule.}
Given a budget of $N$ simulations, MCTS constructs an empirical estimate $\widehat{J}_N(z_{1:K})$
for candidate schedules and returns
\[
z_{1:K}^{(N)} \;\in\; \arg\min_{z_{1:K}\in\mathcal{S}_K \text{ explored}} \widehat{J}_N(z_{1:K}).
\]

We adopt the UCT tree policy \cite{kocsis2006bandit}, which is
asymptotically consistent: under standard assumptions (bounded costs and
unbiased rollout estimates), the empirical estimates
$\widehat{J}_N(z_{1:K})$ converge to the true surrogate cost $J(z_{1:K})$ as
$N \to \infty$. Consequently, the initialization schedule returned by MCTS
converges almost surely to the minimizer of $J(z_{1:K})$ within the feasible
search space.

\begin{theorem}[Bounded dependence minimized by MCTS initialization]\label{thm:mcts}
Under Assumption~\ref{asmp:gap} and the consistency conditions above,
every limit point $z_{1:K}^\star$ of $\{z_{1:K}^{(N)}\}_{N\ge 1}$ satisfies

\begin{align}
  &z_{1:K}^\star \in \arg\min_{z_{1:K}\in\mathcal{S}_K} J(z_{1:K}),
\text{and hence}\nonumber \\
&\sum_{i=1}^K \DepErr_i \;\le\; J(z_{1:K}^\star) \;=\; \min_{z_{1:K}\in\mathcal{S}_K} J(z_{1:K}).  
\end{align}

In particular, among all feasible $K$-step prefixes, the selected initialization attains the smallest achievable \emph{upper bound} on cumulative dependence error.
\end{theorem}

\begin{proof}
By Lemma~\ref{lem:cum}, any schedule $z_{1:K}$ satisfies $\sum_{i=1}^K \DepErr_i \le J(z_{1:K})$.
By the assumed consistency of MCTS, $\widehat{J}_N \to J$ pointwise on explored nodes and the selection rule is asymptotically optimal:
every accumulation point $z_{1:K}^\star$ minimizes $J$ over $\mathcal{S}_K$.
Evaluating the bound at such a minimizer yields
$\sum_{i=1}^K \DepErr_i \le J(z_{1:K}^\star) = \min_{z_{1:K}\in\mathcal{S}_K} J(z_{1:K})$,
which is the tightest bound attainable within $\mathcal{S}_K$.
\end{proof}

\begin{corollary}[Comparison to any baseline prefix]\label{cor:baseline}
Let $\widetilde{z}_{1:K}\in\mathcal{S}_K$ be any baseline schedule (e.g., random or greedy).
Then, under the conditions of Theorem~\ref{thm:mcts}, for large enough $N$,
\[
\sum_{i=1}^K \DepErr_i \;\le\; J(z_{1:K}^{(N)}) \;\le\; J(\widetilde{z}_{1:K}).
\]
Thus the MCTS-chosen prefix achieves a (weakly) smaller upper bound on cumulative dependence than the baseline.
\end{corollary}

\paragraph{Remarks and extensions.}

\begin{remark}[Adding model-error proxies]\label{rem:weights}
If one augments the surrogate with a token-wise model-error proxy
$\mathrm{prox}_\ell$ (e.g., entropy, $1{-}$confidence, top-2 margin),
\begin{align}
    J_\lambda(z_{1:K})
=&
\sum_{i=1}^K \Big[ \lambda_{\mathrm{dep}} B(z_i) + \lambda_{\mathrm{mod}} \sum_{\ell\in z_i} \mathrm{prox}_\ell \Big], 
\end{align}
where $\lambda_{\mathrm{dep}},\lambda_{\mathrm{mod}}>0$, then minimization of $J_\lambda$ with $\lambda_{\mathrm{dep}}>0$ still minimizes
$\sum_i B(z_i)$ subject to the chosen trade-off. The dependence part of Theorem~\ref{thm:mcts} carries through unchanged, yielding the same bound on $\sum_i \DepErr_i$ with $J$ replaced by the dependence component of $J_\lambda$.
\end{remark}

\begin{remark}[On computability and calibration]
All terms in $B(z_i)$ are computed from $p_\theta(\cdot\mid x_{z<i})$ (model logits), so $J$ is observable at inference time.
The tightness of the bound depends on how well the model entropies reflect the coupling structure under $q$; improving calibration may further tighten the practical gap between $J$ and $\sum_i \DepErr_i$.
\end{remark}
\begin{remark}[Feasibility constraints]
The set $\mathcal{S}_K$ can encode architectural or policy constraints (e.g., block sizes, mask connectivity).
Theorem~\ref{thm:mcts} is \emph{relative} to $\mathcal{S}_K$: MCTS finds the tightest upper bound \emph{within} the feasible class.
\end{remark}

\end{document}